\documentclass[sigconf, nonacm]{acmart}

\usepackage{xcolor}
\usepackage{soul,color}
\usepackage{booktabs}
\usepackage{graphicx}

\usepackage{graphics} 
\usepackage{subcaption}
\usepackage{epsfig} 
\usepackage{amsmath} 
\usepackage{comment}

\usepackage{float}
\usepackage{makecell}
\usepackage{graphics} 
\usepackage{amsmath} 
\usepackage[nolist]{acronym}
\usepackage[utf8]{inputenc}
\usepackage{subcaption}
\usepackage{bm}
\usepackage{balance}

\usepackage{xcolor,colortbl}
\definecolor{Gray}{gray}{0.85}
\newcolumntype{C}{>{\centering\let\newline\\\arraybackslash\hspace{0pt}}m{0.1\textwidth}}
\newcolumntype{A}{>{\centering\let\newline\\\arraybackslash\hspace{0pt}}m{0.05\textwidth}}
\newcolumntype{V}{>{\centering\let\newline\\\arraybackslash\hspace{0pt}}m{0.4\textwidth}}
\newcolumntype{L}{>{\let\newline\\\arraybackslash\hspace{0pt}}m{0.85\textwidth}}
\newcolumntype{B}{>{\let\newline\\\arraybackslash\hspace{0pt}}m{0.82\textwidth}}
\newcolumntype{M}{>{\let\newline\\\arraybackslash\hspace{0pt}}m{0.1\textwidth}}
\renewcommand\hl[1]{#1} 

\AtBeginDocument{
}

\copyrightyear{2024}
\acmYear{2024}
\setcopyright{rightsretained}
\acmConference[HRI '24]{Proceedings of the 2024 ACM/IEEE International Conference on Human-Robot Interaction}{March 11--14, 2024}{Boulder, CO, USA} \acmBooktitle{Proceedings of the 2024 ACM/IEEE International Conference on Human-Robot Interaction (HRI '24), March 11--14, 2024, Boulder, CO, USA} \acmDOI{10.1145/3610977.3634948} \acmISBN{979-8-4007-0322-5/24/03}

\makeatletter
\gdef\@copyrightpermission{
  \begin{minipage}{0.3\columnwidth}
   \href{https://creativecommons.org/licenses/by/4.0/}{\includegraphics[width=0.90\textwidth]{images/4ACM-CC-by-88x31.eps}}
  \end{minipage}\hfill
  \begin{minipage}{0.7\columnwidth}
   \href{https://creativecommons.org/licenses/by/4.0/}{This work is licensed under a Creative Commons Attribution International 4.0 License.}
  \end{minipage}
  \vspace{5pt}
}
\makeatother

\begin{document}

\title{``Oh, Sorry, I Think I Interrupted You'': Designing Repair Strategies for Robotic Longitudinal Well-being Coaching}


\author{Minja Axelsson}
\email{minja.axelsson@cl.cam.ac.uk}
\affiliation{%
  \institution{University of Cambridge}
  \country{United Kingdom}
}

\author{Micol Spitale}
\authornote{This work was undertaken and finalised while Micol Spitale was a postdoctoral researcher at the University of Cambridge.}
\email{micol.spitale@polimi.it}
\affiliation{%
  \institution{Politecnico di Milano}
  \country{Italy}
}

\author{Hatice Gunes}
\email{hatice.gunes@cl.cam.ac.uk}
\affiliation{%
  \institution{University of Cambridge}
  \country{United Kingdom}
}


\begin{abstract}

Robotic well-being coaches have been shown to successfully promote people's mental well-being. To provide successful coaching, a robotic coach should have the capability to repair the mistakes it makes.
Past investigations of robot mistakes are limited to game or task-based, one-off and in-lab studies. This paper presents a 4-phase design process to design repair strategies for robotic longitudinal well-being coaching with the involvement of real-world stakeholders: 1) designing repair strategies with a professional well-being coach; 2) a longitudinal study with the involvement of experienced users (i.e., who had already interacted with a robotic coach) to investigate the repair strategies defined in (1); 3) a design workshop with users from the study in (2) to gather their perspectives on the robotic coach's repair strategies; 4) discussing the results obtained in (2) and (3) with the mental well-being professional to reflect on how to design repair strategies for robotic coaching. 
Our results show that users have different expectations for a robotic coach than a human coach, which influences how repair strategies should be designed. We show that different repair strategies (e.g., apologizing, explaining, or repairing empathically) are appropriate in different scenarios, and that preferences for repair strategies change during longitudinal interactions with the robotic coach.
  
\end{abstract}

\ccsdesc[300]{Human-centered computing~User studies}
\ccsdesc[500]{Human-centered computing~HCI design and evaluation methods}
\ccsdesc[500]{Computer systems organization~Robotics}

\keywords{socially assistive robotics, well-being, coaching, interaction ruptures, robot mistakes, human-robot interaction, design research}


\maketitle

\section{Introduction}
Robot mistakes and failures are a very well-known problem in the human-robot interaction (HRI) research community \cite{esterwood2022literature}. For example, robots can hit objects, interrupt people speaking, misunderstand human speech, not respond to people, and experience Wi-Fi malfunctions  \cite{sebo2019don, spitale2023longitudinal}. Therefore, designing repair strategies is extremely important for the success of the human-robot interaction. 
Very recently, the HRI interest in applying and using robots as mental well-being coaches has increased. Robotic coaches for well-being have been examined for use at the workplace \cite{spitale2023robotic, axelsson2023adaptive}, in public environments \cite{axelsson2023robotic, matheus2022social}, in a lab context \cite{axelsson2022participant, churamani2022continual, bodala2021teleoperated}, and for at-home use \cite{jeong2020robotic, jeong2023deploying, jeong2023robotic}. 
In these contexts, addressing the problem of robot mistakes by designing repair strategies is even more relevant for the success of the coaching practice and for making a step forward towards the deployment of such robotic coaches in real-world scenarios.

Past works have investigated the problem of repair strategies \cite{correia2018exploring, van2019take, salem2015would, kontogiorgos2020behavioural} but they have several limitations. First, most of them are limited to game-based (e.g., \cite{sebo2019don}) or task-based scenarios (e.g., \cite{kontogiorgos2020behavioural}), without exploring the extent to which those results may be applicable to other contexts, such as well-being coaching. Second, in most of these studies, the users interacted with the robot in a one-off interaction \cite{sebo2019don, ye2019human}. This constrained the design of repair strategies to short-term interactions without considering longitudinal effects (i.e., how the users' perceptions change over time). Third, most of the studies have been undertaken in the lab \cite{ye2019human, kontogiorgos2020behavioural}, making it difficult to generalise these results to real-world settings.  

In this study, we sought to overcome those limitations by designing repair strategies for robotic longitudinal well-being coaching with the involvement of real-world stakeholders.
To this end, we undertook a 4-phase design process in which we involved a professional mental well-being coach, and users who had already interacted with a robotic coach in their workplace.
As a first step, we designed a set of repair strategies together with the professional human coach, utilizing their expertise and experience. In the second phase, we deployed a robotic coach at a workplace and compared the two sets of repair strategies over four weeks 
by involving 12 users who had already interacted with a robotic coach before. 
After the study, we informed the users of the study set-up, and asked for their feedback on the robot's mistakes in a design workshop.
Finally, we reflected on the user study results and user insights with the professional coach from the first phase, in order to further inform the design of repair strategies for robotic well-being coaches.

This paper contributes to: 1) designing repair strategies appropriate for robotic well-being coaching; 2) investigating the longitudinal effect of repair strategies; and 3) collecting user perspectives on robot mistakes and repair strategies in a real-world context.
%
\begin{figure*}[htb!]
  \centering
  \includegraphics[width = 0.9\textwidth]{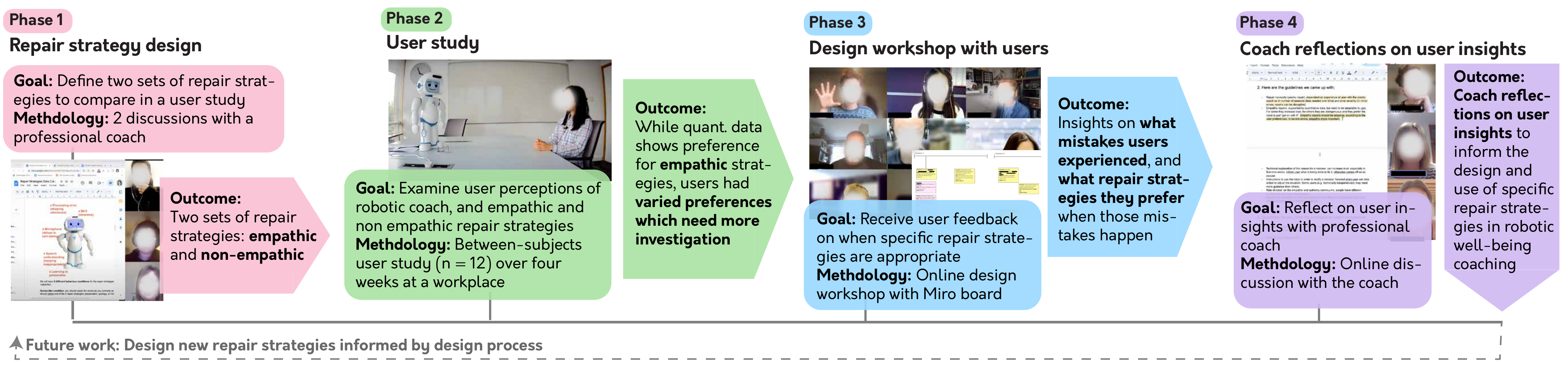}
  \caption{A timeline of the four study phases with their goals, methodologies, and outcomes that steered the next phase.}
  \label{fig:study_phases}
\end{figure*}

\section{Related Work}
\label{sec:related_work}
\textit{Well-being Coaching and Mistakes.}
While there is no single definition of well-being \cite{mcnaught2011defining}, we focus on the definition of well-being as positive psychological functioning \cite{ryff1989happiness}. Mental well-being coaching aims to support coachees to thrive in their life \cite{hart2001coaching} and improve positive psychological functioning \cite{van2023interventions}. Positive psychology coaching in particular aims to help coachees to focus on positive aspects of their life \cite{seligman2007coaching}, through exercises where the coachee e.g. focuses on experiences where they felt grateful, in order to enhance positive memories and goal attainment \cite{wood2010gratitude, emmons2011gratitude}. 
During coaching, difficulties with the coach, such as the coach being vague or not sensitive or supportive enough, not being flexible, and struggling with the concepts of coaching, can disrupt the coach-coachee relationship \cite{carter2017perspectives}. Such issues can negatively impact the \textit{alliance} between the coach and coachee \cite{safran1996resolution}, and as such disrupt the interaction \cite{de2017coaching}. Resolving such mistakes is important so that the coachee can gain the maximum positive impact from coaching. 

\noindent \textit{Robotic Coaching.}
In a workplace coaching context, robotic forms have been compared, finding preference for a toy-like robot \cite{spitale2023robotic}. In public settings, robots have been used to conduct group Mindfulness sessions \cite{axelsson2023robotic}, finding that robots can be helpful but need to be more responsive; and to conduct private deep-breathing sessions at a health centre \cite{matheus2022social}, finding that participants successfully completed deep breathing exercises with the robot. Finally, in home contexts, a robotic coach has been deployed in several at-home studies \cite{jeong2020robotic, jeong2023deploying, jeong2023robotic}, finding that users' well-being improved, and that the robot was perceived more positively when it related to the user as a companion rather than a coach. While these studies all examined robotic coaching in different contexts, none of them focused on the examination and repair of robot mistakes during coaching. Spitale et al. examined how users expressed their behaviour during the mistakes a robotic coach makes \cite{spitale2023longitudinal}, however they did not examine what repair strategies robotic coaches should use when they make mistakes. In this study, we present the first steps into examining \textbf{what repairs are applicable to robotic well-being coach mistakes}.

\noindent \textit{Robot Mistakes and Repairs in HRI.}
Within the context of social interactions, past works have focused on understanding robot failures to improve various aspects of human perception towards robots, such as trust \cite{harrison2023imperfectly}. Correia et al. explored how technical failures of an autonomous social robot affects trust during a HRI collaborative scenario \cite{correia2018exploring}. Their results showed that a faulty robot is perceived as significantly less trustworthy. Analogously, van Wareven et al. investigated the effect of robot failure severity on participants' subjective rating of the robot in a room-escape scenario \cite{van2019take} and found that the severity affects the faith participants had on robots in future scenarios. Salem et al. explored how robot mistakes affect trustworthiness and acceptance in human-robot collaboration \cite{salem2015would}, and showed that subjective perceptions of the robot and assessments of its reliability and trustworthiness has been significantly affected by the robot's performance. Robot mistakes and repair strategies have been largely examined in the context of task-based interactions. Kontogiorgos et al. examined a Furhat robot simulating errors during Wizard-of-Oz cooking instructions \cite{kontogiorgos2020behavioural,kontogiorgos2020embodiment,kontogiorgos2021systematic}, finding that participants responded most positively through non-verbal responses to a robot's explanations. Esterwood and Robert also found that explanations were the most effective strategy for trust repair after repeated mistakes in a box sorting task \cite{esterwood2021you, esterwood2023three}. Additionally, Sebo et al. found that a robot's apology was a more effective trust repair strategy than denial during a game \cite{sebo2019don}. However, these studies did not examine the \textbf{longitudinal effects of different repair strategies}.

\section{Design Process Overview}

This work aims to understand how a robotic longitudinal well-being coach could repair mistakes during coaching. We have undertaken a design process that included four phases depicted at glance in Fig. \ref{fig:study_phases}: (\textbf{Phase 1}) repair strategy design, (\textbf{Phase 2}) user study, (\textbf{Phase 3}) design workshop with users, and (\textbf{Phase 4}) coach reflections on user insights. Each phase was informative for the following one, and we set our goals every step of the way during the design process. 

\textbf{Phase 1} aimed at designing two sets of repair strategies for robotic longitudinal well-being coaches. To accomplish this, we first discussed with a professional well-being coach the robot mistakes in well-being coaching (Discussion 1), and we then defined with them potential repair strategies (Discussion 2), as detailed in Section \ref{sec:phase1}. The results of this design phase were the formulation of empathic and non-empathic repair strategies.
Building upon these results, \textbf{Phase 2} aimed at investigating the user perception towards a robotic coach that used empathic and non-empathic repair strategies in a user study, as described in Section \ref{sec:phase2}. We found that users had various opinions and preferences towards empathic and non-empathic conditions, opening up further investigation. 
Hence, we designed \textbf{Phase 3} as a design workshop with users of the study to better understand the appropriateness of each repair strategy in different scenarios, as detailed in Section \ref{sec:phase3}. The workshop results showed that different repair strategies may be preferred depending on what errors the users experienced, and personal preferences. 
Finally, we involved the mental well-being coach from Phase 1 in \textbf{Phase 4} to gather their feedback and reflect on the findings from Phase 2 and Phase 3, as reported in Section \ref{sec:phase4}.

\section{Phase 1: Repair Strategy Design}
\label{sec:phase1}

In Phase 1, we set out to understand how to design repair strategies for robotic coaches, i.e., what a robot should say when it makes a mistake. To do this, we had two discussions with a professional well-being coach, in order to find out how they repair mistakes during a coaching session, and to design appropriate repair strategies for a robotic well-being coach with the professional coach.

\vspace{2mm}

\noindent \textbf{DISCUSSION 1: Robot Mistakes} 
\label{sec:phase1_discussion1}

\noindent In this first 2-hour discussion with a professional well-being coach, we followed a structure consisting of four parts: 1) show the coach videos of a robotic coach interacting with participants, and instances of interaction ruptures; 2) ask the coach what the robot did wrong; 3) ask the coach how the robot should attempt to repair those situations; and 4) how these repair strategies should best be deployed in a user study. Below, we detail each part of the discussion.

\noindent \textbf{1)} In our previous study \cite{spitale2023robotic}, we collected video recordings of coachees interacting with a robotic coach in a workplace setting, and subsequently analyzed interaction ruptures, i.e. instances where the robotic coach made mistakes and/or where coachees felt \textit{awkward} \cite{spitale2023longitudinal}. We used this video data, and selected short clips that contained instances of interaction ruptures (as annotated in \cite{spitale2023longitudinal}). We showed 6 clips in total, ranging from 22 seconds to 2 minutes and 20 seconds.

\noindent \textbf{2)} The coach noted that the robot's main mistakes were \textit{interrupting} the coachee, and \textit{not providing a response} for a long period of time. Additionally, the robot \textit{did not respond appropriately} to the content of the user's speech. In the videos from our previous study \cite{spitale2023longitudinal}, the robot was responding in a pre-scripted manner, which was a barrier to generating appropriate responses on the go.

\noindent \textbf{3)} The coach mentioned that initially, the robot should \textbf{apologize} for the mistake it has made, and then \textit{identify what mistake} has been made (e.g., ``Oh, sorry, I think I interrupted you.''), as that is what they would do as a coach in a session. Robot apologies have also been previously used to repair trust in human-robot interactions \cite{pompe2022robot}. The coach suggested that after the apology, the robot should then \textit{acknowledge its limitations as a robot}: the robot should \textbf{explain} why a mistake happened, or give its best guess of the technical fault if there is no definitive information (e.g., ``The Wi-Fi is not working well.''); and the robot should emphasize its \textbf{intent to improve} (e.g., ``I'm trying to improve.''), \hl{since self-awareness of robot limitations could put people at ease}. 

\noindent \textbf{4)} In order to examine users' perceptions of a robotic coach's repair strategies, we discussed with the coach whether the robot should make \textit{intentional mistakes} during a coaching session, and then deploy the repair strategies defined above. We decided to examine the most common mistakes of the robot, i.e., interrupting and not responding, \hl{which we identified from the videos}. In order to collect repair strategies with the structure defined above, we scheduled a second discussion with the coach.

\noindent \textbf{Outcome:} Identifying common robot mistakes during coaching interactions (interrupting and not responding), and creating the structure for repair strategies (apology, explanation, and intent to improve).

\vspace{2mm}
\noindent \textbf{DISCUSSION 2: Repair Strategies}
\label{sec:disc2}

\noindent In this 1.5-hour discussion, we used \textit{bodystorming} \cite{oulasvirta2003understanding}, i.e., we simulated an interaction with the robotic coach by asking the professional coach to act as the robotic coach, and two researchers acting as coachees. We did this in order to experience the coaching session, the mistakes, and the perception of the planned repair strategies from the perspective of the coachees. In preparation for the discussion, we adapted four existing well-being exercises with the coach: (1) \textit{savouring}, where the coach asks the coachee to reflect on a positive memory in the recent past (\cite{smith2019effects}); (2) \textit{gratitude}, where the coach asks the coachee to think of things they were grateful for (\cite{gregersen2014examining}); (3) \textit{accomplishments}, where coachees reflect on accomplishments (\cite{gregersen2014examining}); and (4) \textit{one door closes one door opens}, where coachees think of a time when a door closed (i.e., they missed out on an opportunity), and what doors opened as a result (i.e., new opportunities arose), in order to cultivate optimism (\cite{kauffman2006positive}). 

During the bodystorming, we asked the coach to act as a robot and intentionally interrupt and not respond to the researchers, and then use the repair strategy we designed (see Sec. \ref{sec:phase1_discussion1}) in Discussion 1 (i.e., apology, explanation, and intent to improve). We provided the coach with a list of causes for robot errors such as interrupting and not responding, to use in their \textbf{explanation}: microphone fault, processing error, slow Wi-Fi, and error in speech understanding.

We structured the 1.5-hour session as follows: 1) 30 minutes of coaching, mistakes and repair strategies with 15 minutes per each researcher while the other took notes; 2) 15 minutes of discussion; 3) 30 minutes of coaching, mistakes and repair strategies with 15 minutes per researcher while the other took notes; and 4) 15 minutes of discussion. We detail each part of the discussion below.

\noindent \textbf{1)} The coach conducted each exercise with the two researchers, acting as the robotic coach, and asked for 2 instances of positive experiences during each exercise (e.g., 2 moments when the coachee felt grateful). During each instance, the coach made a mistake (not responding or interrupting), and then used a repair strategy (apology, explanation, and intent to improve). We collected 8 different phrasings of repair strategies in this manner.

\noindent \textbf{2)} After the first half hour, both researchers and the coach had a discussion on how the session went. Both researchers noted that despite the employed repair strategies, they \textit{did not feel understood or listened to}, and that the coach \textit{did not understand how they felt}. The coach also mentioned that it could be helpful to \textit{ask the coachee how they felt} after the coach made a mistake. Both researchers noted that 
the coaching interaction appeared \textit{awkward}. We decided that to resolve these issues (which have also been previously reported in robotic coaching \cite{axelsson2022participant, spitale2023robotic, axelsson2023robotic}), the robotic coach should be \textbf{empathic}. In fact, empathic communication in a therapeutic context can help a client feel listened to, understood and accepted, and have their feelings validated---improving outcomes for well-being \cite{jani2012role, watson2014role}. Empathy can also be conducive to resolving affective ruptures in interpersonal interactions \cite{galinsky2011using}. The coach suggested that we amend the repair strategy structure to be empathic as follows (additions in bold): apology, \textbf{ask for user emotion}, \textbf{empathize with user emotion}, \textbf{reassure user}, explanation, and intent to improve.

\noindent \textbf{3)} For the second 30-minute coaching session, each researcher again did two exercises with the coach. 
This time, the coach used the \textbf{empathic} repair strategies as designed above. Again, the coach administered each of the four exercises, with 2 instances of each, resulting in 8 different phrasings of empathic repair strategies.

\noindent \textbf{4)} In the final discussion, we found that the coach felt the empathic strategies better suited a robotic coach, and the researchers felt \textit{listened to and heard}, and that the coach \textit{understood how they felt}. The researchers also observed that the interaction appeared less \textit{awkward}. We concluded that in order to examine how \textbf{empathy impacts users' perceptions of repair strategies}, we should conduct a \textit{between-subjects study} comparing the two different types of repair strategies (empathic and non-empathic). \hl{We also discussed that in order to investigate how users' opinions of robotic coach repair strategies evolve over time, the study should be longitudinal.}

\noindent \textbf{Outcome:} Study design of comparing empathic and non-empathic repair strategies, and 8 phrasings of each type of repair strategy.



\section{Phase 2: User Study}
\label{sec:phase2}

In order to examine the empathic and non-empathic repair strategies defined in Phase 1 (Sec. \ref{sec:phase1}), we conducted a longitudinal user study where the robot executed those repair strategies (Phase 2 in Fig. \ref{fig:study_phases}). The study design, the experiment protocol, and the consent forms were approved by the Ethics Committee of the Department of Computer Science and Technology, University of Cambridge.

\subsection{Protocol \& Questionnaires}
\label{sec:protocol_and_questionnaires}

We conducted the study over four weeks, with one well-being exercise (a maximum of 10 minutes) administered by the robotic well-being coach per week. Users interacted with the robot at their workplace, in a room reserved for the study, with the robot standing on a table 1.5 meters away from the user, and the user sat at a chair next to the table. 
After each interaction, users filled in a PANAS questionnaire about their positive and negative emotions \cite{watson1988development}, RoPE questionnaire about the user's perception of the robot's empathy \cite{charrier2019rope}, RoSAS about the user's perception of the robot's social attributes \cite{carpinella2017robotic}, and the MDMT questionnaire about trust in the robot \cite{malle2021multidimensional}. 
At the end of the study, users took part in a semi-structured interview about their overall experience with the robot (15-20 minutes). This interview had two parts: one with general questions about the robot, its mistakes and repair strategies; and one after disclosing the study protocol and the pre-planned mistakes to the user, followed by questions about the robot's mistakes and repair strategies. Interview questions are reported in the Supplementary Material (Sec. 1).

\subsection{Users}
\label{sec:users}

The users ($n = 12$) were recruited from the host company called Cambridge Consultants Inc., and had previously interacted with a robotic well-being coach in a 4-week study \cite{spitale2023robotic}, where they experienced commonly known robot errors (such as interruptions and slow responses). We selected these experienced users in order to to mitigate the \textit{novelty effect} \cite{belpaeme2020advice}, so that we could engage users in a critical discussion about the robot's mistakes and repair strategies after the study, which would not be influenced by the novelty of a robotic well-being coach. Users were also screened for anxiety (GAD-7 questionnaire) \cite{williams2014gad} and depression (PHQ-9 questionnaire) \cite{kroenke2001phq}. We chose this screening in order to not use a robotic coach with a clinical population, which we do not consider ethical prior to thorough examination with a non-clinical population. The users were split into two groups ($n = 6$ per condition), one group experiencing the \textbf{empathic} repair strategy condition, and the other the \textbf{non-empathic} repair strategy condition. 1 user was aged 18--25, 4 were aged 26--35, 3 were aged 36--45, and 4 were aged 46--55. 3 users were female, 1 non-binary, and 8 male. Minority genders (female and non-binary) were balanced across conditions, with 2 in each condition. Users rated their previous experience with social robots as $(M = 3, SD = 1.044)$ on a scale from 1 (lowest) to 5 (highest). All users were native or fluent speakers of English, and all had an Undergraduate, Master's, or PhD degree. The users' demographics reflect the demographics of the host company Cambridge Consultants Inc., and as such have the distribution of the ages, genders, and degree statuses as described.

\subsection{Robot Platform and Architecture}

We used the QTrobot by LuxAI S.p.A.\footnote{https://luxai.com/}---a 90 cm tall, tabletop child-like robot with static legs, 4 degrees of freedom (DOF) arms, 2 DOF neck, and a screen face. We chose this robot as it has been previously used successfully as a robotic coach \cite{spitale2023robotic, axelsson2023robotic}. The fully autonomous robotic coach was implemented using our newly developed VITA system \cite{spitale2023vita}, a multi-modal LLM-based system for longitudinal and adaptive robotic mental well-being coaching, that is open source 
leveraging on the HARMONI framework \cite{spitale2021composing}.

\subsection{Exercises}
\label{sec:exercises}

The robotic coach administered a different \textit{Positive Psychology} exercise each week, one per week. The robotic coach delivered the same four exercises used during Discussion 2 described in Section \ref{sec:disc2}. Each exercise consisted of the robot asking for two different examples from the users, and two follow-up questions per example. Follow-up questions were generated by sending the user's utterance in response to the robot's questions to ChatGPT \textit{gpt-3.5-turbo} model via OpenAI APIs\footnote{\label{openai}https://platform.openai.com/}. We used ChatGPT to generate follow-up questions, in order to minimize the impact of a pre-scripted robot not responding appropriately, which was a mistake the coach identified from the robotic coach videos in Phase 1 (Sec. \ref{sec:phase1_discussion1}). 

\subsection{Administered Robot Mistakes and Repairs}

Each week, the robot was pre-programmed to make two mistakes, each during one of its utterance turns during the interaction. \hl{While the conversational flow itself was automated (as described in Sec.} \ref{sec:exercises}), \hl{this conversational flow was interrupted at pre-determined times to administer the mistake and the repair.} The timing and type of mistake was counterbalanced across the sessions to avoid repetitiveness (each timing and type is listed in Tables 1, 2 and 3 in the Supplementary Material). These mistakes were either interrupting the user (3-7 seconds into the user's speaking turn), or not responding to the user for a longer period of time (12-18 seconds). These mistakes were chosen since they were the most common mistakes in robotic well-being coaching, as identified from watching the robotic coaching videos with the professional coach in Phase 1 (Sec. \ref{sec:phase1_discussion1}). 
The robot made these mistakes at different stages (either after initially explaining the exercise, or when asking the follow-up questions). The mistakes were distributed as follows: (Session 1) interrupting and not responding, (Session 2) interrupting twice, (Session 3) not responding twice, (Session 4) not responding and interrupting. 

\subsection{Study Conditions}

Our two study conditions for the between-subjects study were the robot administering either \textbf{empathic} repair strategies, or \textbf{non-empathic} repair strategies. The repair strategies were deployed whenever the robot made a pre-planned mistake, and were constructed together with the professional well-being coach. 

The basic structure of both conditions was defined together with the professional well-being coach, to include an apology, a technical explanation for why the error occurred 
(realistic explanations for why each type of mistake typically occurs in HRI---e.g., microphone or Wi-Fi malfunction \cite{spitale2023longitudinal}), and intent to improve (to reassure the user). In the \textbf{empathic} condition, the robot would also ask the user about the emotion they were experiencing due to the mistake, cognitively empathize with the emotion \cite{birmingham2022perceptions} (i.e. repeating the emotion of the user to acknowledge it), and affectively reassure them \cite{pincus2013cognitive} (aiming to reduce worry and to reassure the user that the robot is attempting to listen to them). To cognitively empathize and affectively reassure the user, the user's utterance in response to the robot's question about how they were feeling was sent to ChatGPT \textit{gpt-3.5-turbo} model via OpenAI APIs
, and ChatGPT was prompted to return the utterance's \textbf{emotional valence} (positive, neutral, or negative), and the \textbf{specific emotion} (repeated back to the user by the robot when ChatGPT returned the utterance's emotion as negative, to acknowledge it).


\subsubsection{Repair Strategy Construction} 

The repair strategies and examples of each were constructed (with differences between conditions italicized) with the following structure. The specific wording of each repair strategy was different, and was based on the professional coach's phrasing. A full list of all repair strategies is made available in Supplementary Material (Sec. 2).

\noindent \textbf{Non-empathic:} Apology, explain technical error, intent to improve, ask for repetition of user's previous utterance

\noindent \textbf{Example: } ``Oh, sorry, I think I interrupted you. My microphone isn't working well today. I'm trying to do better. Could you repeat what you were saying before I interrupted you?''

\noindent \textbf{Empathic:} Apology, \textit{ask for user emotion}, \textit{cognitively empathize with user emotion}\cite{birmingham2022perceptions}, \textit{affective reassurance}\cite{pincus2013cognitive}, explain technical error, intent to improve, ask for repetition of user's previous utterance.

\noindent \textbf{Example: } ``Oh, sorry, I think I interrupted you. How did me interrupting you make you feel?'' 
 
    $\rightarrow$ \textbf{Negative user feeling description: } \newline
    \indent ``\textit{I'm sorry, I understand it can make you feel <feeling [e.g., awkward]> when I make mistakes. My intention is to listen to what you are saying, but sometimes I experience errors.} My microphone isn't working well today. I'm trying to  do better. Could you repeat what you were saying before I interrupted you?''

    $\rightarrow$ \textbf{Positive or neutral user feeling description: } \newline
    \indent ``\textit{Thanks for being understanding. My intention is to listen to what you are saying, but sometimes I experience errors.} My microphone isn't working well today. I'm trying to do better. Could you repeat what you were saying before I interrupted you?''


\subsection{User Study Findings}
\label{sec:findings_userstudy}

\subsubsection{Data Analysis}
Due to the sample size for a between-subjects study
, we use \textit{descriptive statistics} to describe the quantitative differences between user groups' perceptions of the empathic and non-empathic repair strategies, as well as the longitudinal perception of the repair strategies. For qualitative analysis, we use Framework Analysis \cite{srivastava2009framework}, consisting of the steps of: (1) familiarization with the data, (2) identifying a thematic framework, (3) indexing, (4) charting the data, and (5) interpretation of the data.

\subsubsection{Quantitative Results}

We present these results to contextualize our qualitative results, as well as the user design workshop (Sec. \ref{sec:phase3}) and coach feedback (Sec. \ref{sec:phase4}) findings. As described in Sec. \ref{sec:protocol_and_questionnaires}, we administered questionnaires after each session, and after the four sessions of the study. We also measured users' well-being with Ryff's well-being questionnaire \cite{ryff1989happiness} pre- and post-study (min.: 18, max.: 108). The median of well-being was Empathic: ($(pre = 100, post = 105)$); Non-empathic: ($(pre = 90.5, post = 92.5)$). These results confirmed our expectations of no significant impact on well-being, due the negative impact of the planned mistakes on the coaching experience, as well as the screened user group with high levels of well-being to begin with (see Sec. \ref{sec:users}). 

\begin{table}[]
\resizebox{\columnwidth}{!}{%
\begin{tabular}{@{}lllll@{}}
\toprule
\textbf{Questionnaire} & \textbf{Measure} & \textbf{Empathic} & \textbf{Non-Empathic} & \textbf{Average} \\ \midrule
         \textbf{WAI-SR \cite{munder2010working}}  &  Goal-subscale $\uparrow$ & \textbf{M $=$ 7.500, SD $=$ 1.225} & $(M = 5.667, SD = 1.506)$ & $(M = 6.583, SD = 1.621)$ \\
         \textbf{WAI-SR \cite{munder2010working}}  &  Task-subscale $\uparrow$ & \textbf{M $=$ 14.000, SD $=$ 2.098} & $(M = 8.167, SD = 2.714)$ & $(M = 11.083, SD = 3.825)$ \\
         \textbf{WAI-SR \cite{munder2010working}}  &  Bond-subscale $\uparrow$ & \textbf{M $=$ 31.333, SD $=$ 4.761} & $(M = 21.333, SD = 7.340)$ & $(M = 26.333, SD = 7.878)$ \\
         \textbf{WAI-SR \cite{munder2010working}}  &  Alliance total $\uparrow$ & \textbf{M $=$ 31.333, SD $=$ 4.761} & $(M = 21.333, SD = 7.340)$ & $(M = 26.333, SD = 7.878)$ \\ 
         \textbf{SUS \cite{grier2013system}}  & Usability total $\uparrow$ & \textbf{M $=$ 3.500, SD $=$ 1.517} & $(M = 2.500, SD = 0.837)$ & $(M = 3.000, SD = 1.280)$ \\ 
         \textbf{(C) - Understand}  &  What I said $\uparrow$ & \textbf{M $=$ 3.500, SD $=$ 1.517} & $(M = 2.500, SD = 0.837)$ & $(M = 3.000, SD = 1.280)$ \\  
         \textbf{(C) - Understand}  &  How I felt $\uparrow$ & \textbf{M $=$ 2.500, SD $=$ 1.643} & $(M = 1.500, SD = 0.837)$ & $(M = 2.000, SD = 1.349)$ \\  
         \textbf{(C) - Understand}  &  Adapted $\uparrow$ & \textbf{M $=$ 3.500, SD $=$ 1.055} & $(M = 3.000, SD = 0.894)$ & $(M = 3.25, SD = 1.055)$ \\  
         \textbf{(C) - Mistakes}  &  Made mistakes $\uparrow$ & \textbf{M $=$ 4.8.333, SD $=$ 0.408} & $(M = 4.500, SD = 0.837)$ & $(M = 4.667, SD = 0.651)$ \\  
         \textbf{(C) - Mistakes}  &  I understood why $\uparrow$ & \textbf{M $=$ 4.167, SD $=$ 1.170} & $(M = 3.667, SD = 1.033)$ & $(M = 3.917, SD = 1.084)$ \\  
         \textbf{(C) - Mistakes}  &  Irritation $\downarrow$ & $(M = 3.167, SD = 1.329)$ & \textbf{M $=$ 4.167, SD $=$ 1.169} & $(M = 3.667, SD = 1.303)$ \\  
         \textbf{(C) - Mistakes}  &  Disruption $\downarrow$ & $(M = 3.667, SD = 1.506)$ & \textbf{M $=$ 4.500, SD $=$ 0.837} & $(M = 4.083, SD = 1.240)$ \\  
         \textbf{(C) - Mistakes}  &  Repaired -- & \textbf{M $=$ 3.167, SD $=$ 1.412} & \textbf{M $=$ 3.167, SD $=$ 0.983} & \textbf{M $=$ 3.167, SD $=$ 1.193} \\  
         \textbf{(C) - Repairs}  &  Appropriate $\uparrow$ & \textbf{M $=$ 3.333, SD $=$ 1.211} & $(M = 3.000, SD = 1.095)$ & $(M = 3.167, SD = 1.115)$ \\  
         \textbf{(C) - Repairs}  &  Appropriate amount $\uparrow$ & \textbf{M $=$ 3.333, SD $=$ 1.033} & $(M = 2.833, SD = 0.753)$ & $(M = 3.083, SD = 0.900)$ \\  
         \textbf{(C) - Repairs}  &  Right time $\uparrow$ & \textbf{M $=$ 2.833, SD $=$ 1.170} & $(M = 2.500, SD = 1.049)$ & $(M = 2.667, SD = 1.073)$ \\  
         \textbf{(C) - Repairs}  &  Empathic $\uparrow$ & \textbf{M $=$ 3.667, SD $=$ 1.366} & $(M = 2.833, SD = 1.330)$ & $(M = 3.250, SD = 1.357)$ \\  
         \bottomrule
\end{tabular}
}
\caption{Post-study quantitative measures (higher measures bolded). (C) denotes a custom question. Arrows illustrate where positive changes occurred in the empathic condition.}
\label{table:quant_endofstudy}
\end{table}

We present post-study quantitative measures in the Table \ref{table:quant_endofstudy}. As post-study measures, we administered the WAI-SR (Working Alliance Inventory Short) questionnaire \cite{munder2010working} to measure user alliance with the robotic coach, and the SUS (System Usability Scale) questionnaire \cite{grier2013system} in order to examine users' perceptions of the robotic coach between conditions. Additionally, we administered custom questions on a Likert scale from 1 (lowest) to 5 (highest), about the robot's understanding of what the user \textit{said}, \textit{felt}, and \textit{how it adapted to them}; the robot's mistakes in terms of whether it \textit{made mistakes}, whether the users \textit{understood why} it made mistakes, were \textit{irritated} by the mistakes, the session was \textit{disrupted} by the mistakes, and whether the robot \textit{repaired} the mistakes; and on the repairs in terms of how \textit{appropriate} they were, how appropriate the \textit{amount} of repairs was, whether the repairs were administered at the \textit{right time}, and whether the repairs were \textit{empathic}. The data indicates that the robot was perceived slightly more positively in the \textbf{empathic} condition for alliance, usability, understanding, and success of repairs than in the non-empathic condition. For mistakes, the robot was also perceived more positively, with users better understanding why the robot made mistakes, and feeling less disrupted and irritated.


\begin{figure}[]
  \centering
  \includegraphics[width = 0.85\columnwidth]{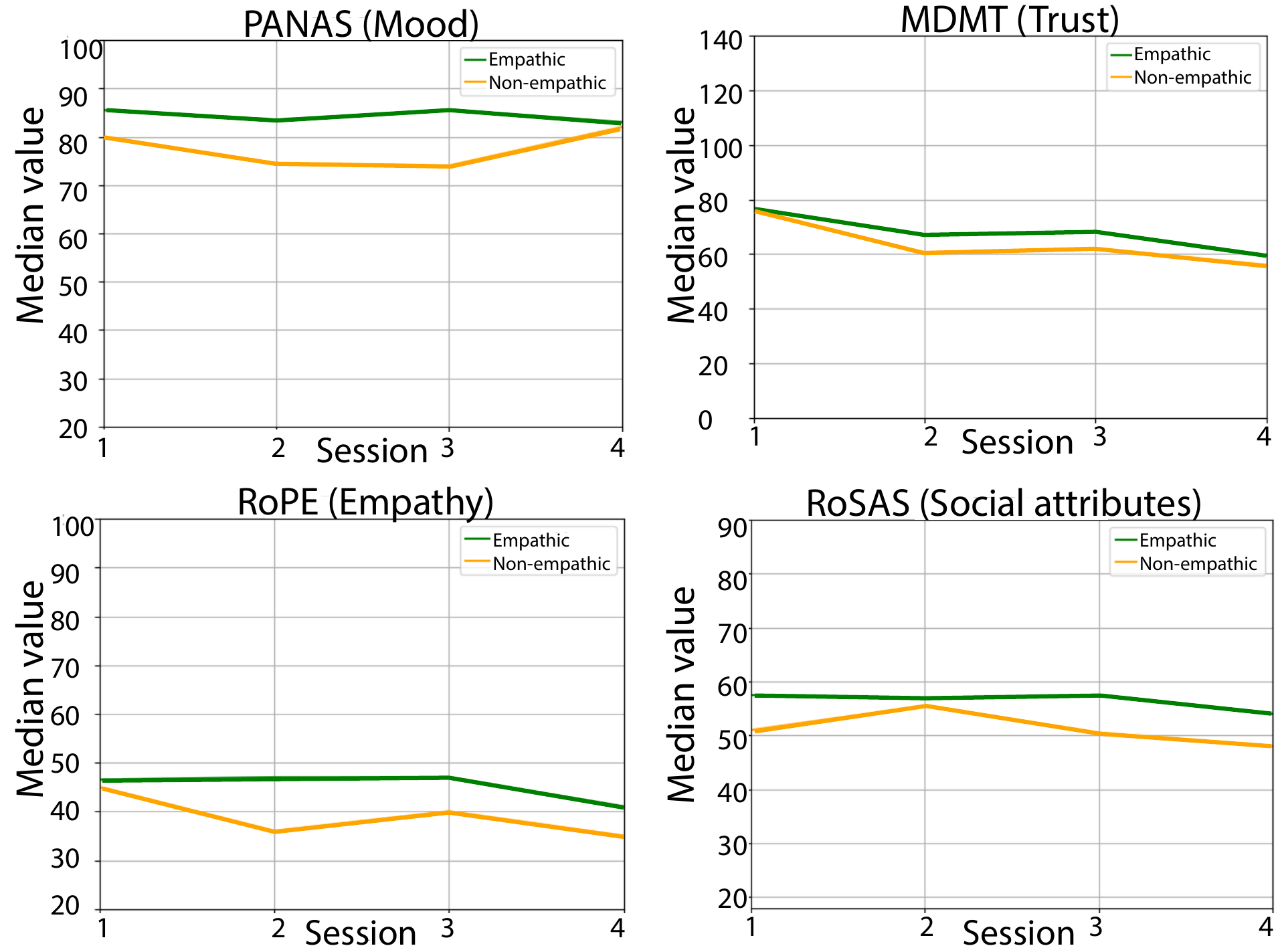}
  \caption{Longitudinal trends of quantitative measures in users' experiences of \textit{empathic} and \textit{non-empathic} repairs.}
  \label{fig:longitudinal_results}
\end{figure}

We find the longitudinal quantitative data from week-by-week measures to be a useful indicator of users' experience. We calculated the median for each measure (reported in Sec. \ref{sec:protocol_and_questionnaires}) for each week, within each condition group and across all users. Our data show decreasing trends across all users and within both conditions throughout the weeks (Fig. \ref{fig:longitudinal_results}) for trust in the robot (MDMT), robot's perceived empathy (RoPE), and a similar trend for the empathic condition for the robot's social attributes (RoSAS). These data indicate that over time, the users' experience with the robot was decreasing in quality. The qualitative data in the next section (Sec. \ref{sec:user-study_qual-results}) helps understand why this may be the case. Mood (PANAS) measures remained relatively stable across the weeks, with the non-empathic group experiencing lower mood throughout. Together with the other measures, this may indicate a more negative experience with the robot in the non-empathic condition. 

%



\subsubsection{Qualitative Results}
\label{sec:user-study_qual-results}
We analysed our qualitative results with Framework Analysis \cite{srivastava2009framework}, to examine users' longitudinal experiences, and the two condition groups (empathic and non-empathic). We present quotes related to these themes in the Supplementary Material (Sec. 3). For instance, some users called the empathic strategies of the robot ``caring'' (P20) and that it ``adapted to the sentiment'' they were feeling (P01), but that the empathy became less genuine over time (P01) and was ``a bit over the top'' (P05). Some users stated a preference for not needing empathy in a robot: ``it felt too much like a machine and empathy doesn't apply'' (P02).

In terms of longitudinal experiences, users noted that in the initial weeks, they were trusting of the robot recognizing its mistakes, and of its repairs. However, this trust decreased throughout the weeks, as shown by quantitative (Fig. \ref{fig:longitudinal_results}). For instance, explanations were viewed as helpful to understand the robot was not ``broken'' (P09), however users noted that the robot's explanations seemed \textit{less genuine} towards the end of the study and felt like ``excuses'' (P11). In fact, some users noted that repairs ``made [the interaction] worse'' (P04), disrupting the interaction. The repetitiveness of the repairs was also viewed as disruptive, and that they were not ``meeting my needs, as in not understanding'' (P06). 

The challenges for repair strategy design found by this analysis are 1) users become less receptive to the repair strategies over time, 2) users' personal preferences for repairs (especially empathic vs non-empathic repairs), and 3) repairs are sometimes helpful but sometimes disruptive. 
Users had extensive feedback on how repair strategies could be improved, and when and how often to apply them.
To investigate this further, we organized a design workshop with the users as the next phase of the study. 





\section{Phase 3: Workshop with Users}
\label{sec:phase3}

We conducted a design workshop with users from the study in Phase 2 ($n = 10$, with P04 and P07 not available to attend), in order to gather their opinions on how a robotic well-being coach should deploy repair strategies in different scenarios during coaching. We discussed the challenges identified in Phase 2 (Sec. \ref{sec:phase2}), namely the negative perception of repair strategies over time, users' personal preferences, and when repairs are helpful and when disruptive. The workshop was conducted in an online video call, with a Miro board for collaboration\footnote{https://miro.com/}. 
The workshop was structured into five sections: (1) showing users videos of both repair strategy conditions (empathic and non-empathic) to inform them of the robot's capabilities, (2) warm-up exercise
, (3) thinking through appropriate repairs for pre-defined scenarios, and generating mistake scenarios that each user experienced, (4) thinking through appropriate repairs for those user-generated scenarios, and (5) distilling insights for robotic well-being coach repairs for different contexts. 

\subsection{Repair Strategy Preferences for User-generated Mistake Scenarios}

\label{sec:user-generated_mistake_scenarios}

We have included the 12 user-generated mistake scenarios in the Supplementary Material (Sec. 4). In addition to the robot's pre-programmed mistakes of interrupting and non-responding, users generated other scenarios (referred to as ``Sc.'' throughout this paper) of robot mistakes, e.g. the robot misunderstanding them or asking generic questions. Users could select multiple options for repair strategies for their scenarios, from 5 categories \hl{(including the categories ``empathic'' and ``technical explanation'' which they experienced in the study, and ``instructions on use'', ``brief apology'', and ``do nothing'', which they suggested in the post-study interviews)} and also add another suggestion in the ``other'' section. Overall, for the 12 user-generated scenarios, user responses from the categories tallied up as follows: \textit{do nothing} (32), \textit{empathic} (15), \textit{brief apology} (48), \textit{technical explanation} (6), and \textit{instructions on use} (10).

\noindent \textbf{Do nothing or apologize briefly --} Users show an overall preference for the robot either \textit{doing nothing} or giving a \textit{brief apology.} Reasons for these included ``For a small error just continue with the session'' (P06, Sc. 4), ``Better to move on'' (P01, Sc. 4), ``Not worth dwelling over, but [use] quick apology for throwing the user off.'' (P08, Sc. 5). Users explained that longer repairs may in some cases distract from the session, and the robot should ``continue momentum'' (P01, Sc. 4) and ``try and get [the] conversation back on track'' (P05, Sc. 5). 

\noindent \textbf{Empathic --} There were no scenarios where most users wanted empathic repairs. In scenarios where users did want them (2-3 users), the reasons were e.g. ``Being empathetic, efficient and give advice on improving user experience seems useful.'' (P11, Sc. 2), 
and ``
Any empathetic apology should be to validate user feelings and be quick [..]
Providing advice on how to avoid this error would be useful again to continue conversation along.'' (P11, Sc. 7). One user noted that a repair ``Should be an empathic apology but not necessarily asking how user felt.'' (P01, Sc.10), wanting an empathic repair but in a brief format.

\noindent \textbf{Technical explanation and Instructions on use --} Users preferred \textit{instructions} over \textit{explanations.} Reasons for instructions were e.g. ``I would like to know how to get robot to repeat question in the future if it is my fault.'' (P11, Sc. 9), ``If the robot can't adapt to the user can the users adapt to the robot'' (P06, Sc. 10). Users wanted instructions to interact with the robot when they could be able to correct the error. In contrast, explanations should be used e.g. ``If the reason for not understanding is known, try and point it out to the user so they can address it.'' (P05, Sc. 10) and ``Given it's a multiple repeat I'd want to know that it has realised it's mistake and by giving an explanation I would assume it can actually move on.'' (P03, Sc. 11). In these cases, the user wanted the robot to inform them of the cause of the error to increase the transparency of whether the robot is aware of the error and the reason for it, as well as whether the user should further address it (outside of the session). Technical explanations and instructions could be used in 
tandem, e.g., ``Robot should explain the technical error, perhaps it did not hear properly and microphone should be moved closer.'' (P09, Sc. 7).


\subsection{User Insights and Discussion}
\label{sec:user_insights}

While the previous sections show that users had significant \textit{personal preferences}, we conducted a final discussion at the end of the design workshop to shape some more general insights for robotic coach repair strategies. These insights are detailed here.

\noindent \textbf{To repair or not to repair? --} \textbf{In general, users wanted the robot to repair when the mistake was more disruptive to the session than deploying a repair would be.} \hl{Users noted the robot \textit{should not repair} minor mistakes, because paying attention to the mistake could cause further disruption to the session. Minor mistakes were e.g.}, ``if it interrupts the user momentarily'' (P11), ``it interrupts the user and the user keeps talking'' (P08), ``if it's likely a one-off issue that will not impact further'' (P03). In these cases, the users preferred the robot to either \textit{do nothing} or \textit{give a brief apology}. The robot should also not repair if it disrupts the session: ``if the user is in the flow and it would be detrimental to the session to interrupt the user'' (P10), and ``the repair would be more interrupting / distracting than what is trying to be repaired''. Users noted that the repairs would need to be selected to be less disruptive (i.e., in the case of a minor mistake, a more detailed repair than a brief apology might be disruptive). Scenarios where the robot \textit{should definitely repair} were ``when the user needs to participate in the fix'' (P03), or ``if the robot's response depends on the user'' (P11). \hl{In terms of longitudinally administering repairs}, \hl{users noted that initially the robot should focus on introducing its main functionalities. In the next few sessions, it is important to administer repairs to introduce the repair capability to the user and improve their experience (P01), and to give the user any necessary instructions to resolve issues (P10). Users noted that over time they would expect less repairs, since the robot is not a ``stranger'' anymore (P01), and that it ``can cut through and carry on'' (P10). } 

\noindent \textbf{When to use empathic repairs? --} \textbf{In general, users wanted empathic repairs when they were likely to feel frustrated}. This could be the case in ``repeated mistakes, especially when I have been talking for a while already'' (P05), ``when it interrupts during a story/detailed explanation of an experience'' (P03), and ``when the error has affected the session in a way that could have negatively impacted the flow'' (P10). However, one user noted that empathic repairs could make the user ``feel more frustrated'' (P01). This indicates that empathic repairs should be dynamically deployed based on the user's response to the repair itself.

\noindent \textbf{When to give technical explanations? --} \textbf{In general, explanations were wanted to increase transparency} in the case of severe technical errors. For example, ``when the error has occurred for the first time overall for rare errors'' (P12), and ``when an error has occurred that is outside the user's control'' (P10). However, the explanation should be ``something the user can understand'' (P03).

\noindent \textbf{When to give instructions on use? --} \textbf{In general, users wanted instructions from the robot when the mistake was something that the user can help correct,} during the first few interaction sessions with the robot. Examples were ``when the user can actually make a simple fix (like move the microphone closer)'' (P03) and ``if the error has to do with the user's way of communicating'' (P11). 


\section{Phase 4: Coach Reflections}
\label{sec:phase4}

To reflect on the insights provided by the robotic coach users, we took these insights back to the professional well-being coach who had collaborated with us to design the robotic coach and its repair strategies in Phase 1 (Sec. \ref{sec:phase1}). In this discussion (1 hour), we asked the coach to reflect on the results of our user study and the collected user insights, and to give their opinions on them, as follows. 


\noindent \textbf{Robotic coach repair strategy design can not be solely based on professional coach repair strategies} --- We can not conclude simply that ``empathic repair strategies are better than non-empathic ones'', as was the \hl{original} expectation of the authors and the well-being coach. Instead, our findings indicate that a \textit{robotic well-being coach is fundamentally different in its capabilities when compared to a human coach, and as such users' expectations are different}. \hl{The professional coach noted that repeated social mistakes (interrupting or non-responding) in human coaching are rare, as humans have the capacity to better intuit social signals and thus make fewer mistakes. As coachees do not expect many mistakes in human coaching, empathic repairs may be helpful in the rare situations where mistakes do occur. 
However, due to limited robot capabilities, social mistakes occur often in robotic coaching (as shown by the videos in Sec.} \ref{sec:phase1} \cite{spitale2023longitudinal}). \hl{Due to repeated mistakes, while the coach's and researchers' intuition was that empathic repairs would be similarly applicable to a robotic coach, the user study (Sec.} \ref{sec:phase2}) \hl{and design workshop with users} (Sec. \ref{sec:phase3}) \hl{contradicted this intuition. Some users enjoyed empathic repairs in the first few interaction sessions, but became averse when repeated in later interactions, viewing them as disingenuous. Other users experienced the empathic repairs as disingenuous already from initial interactions, due to viewing the robot as a tool that empathic expressions are incongruous with. As such, repair design needs to acknowledge the limited capabilities and higher error rate of robotic coaches.} 


\noindent \textbf{Repair strategies are not always necessary} --  In some instances users perceived the repair strategies as disruptive (see Sec. \ref{sec:phase2}-\ref{sec:phase3}). 
The professional coach agreed that repairs could be detrimental if they were ``long-winded explanations'' that may be perceived as excuses rather than helpful. The coach agreed that users may ``get used to a robot and its errors'', and due to this, repairs could be reduced over time in the case of repetitive errors. \textbf{Future research} is needed in to distinguish whether and when a user no longer requires repair for a specific error, e.g. by detecting a user's behavioral signals.

\noindent \textbf{Repairs should be utilized according to user preference} -- There is no ``one-size-fits-all'' with regards to non-/empathic (or other) repairs in robotic well-being coaching. 
Some users appreciated the empathic repairs and viewed them as increasing trust and improving the interaction, while some viewed them as disingenuous and even deceptive due to the fundamentally non-empathic nature of a robot as a machine. The professional coach noted that at times they adjust their level of empathic expression to their client, by matching their expression to that observed from the client. 
Previous research proposed the adaptation of a robot's empathic expressions to a user's positive emotional signals \cite{leite2012modelling}. \textbf{Future research} is needed to analyze user feedback during the administration of the repair strategy itself (i.e., observing a user's response to a repair via behavioral signals), and accordingly adjusting future repairs.

\noindent \textbf{Robot-specific repair strategies include technical explanations and instructions on use} -- The professional coach noted that technical error explanation is a robot specific behaviour, and there is no direct comparison to human-to-human coaching, but that if they were making a repeated error, they would want to give an explanation to their client. \textbf{Future research} is needed to develop robot awareness of mistake occurrence, cause, and to generate authentic technical explanations. In terms of instructions on use, the coach compared this to the situation where their client may have hearing loss, and the coach might in turn increase their speech volume. \textbf{Future research} should focus on how robots may detect and adapt to such personal requirements of each user, while applying appropriate personalised instructions on use of the robot.

\noindent \hl{\textbf{Repetitions of repair strategies are detrimental} -- Due to social interaction timing issues (i.e., mistakes such as interruptions and non-responding) that are inherent to social robots} \cite{spitale2023longitudinal, maitreyee2021younger, honig2018understanding}, \hl{robotic coaching can have an awkward rhythm. When repair strategies are continuously administered to repair these issues (even with different phrasings across sessions, as we have done in our user study), the effect can be detrimental to the user. The coach compared this to a ``phone helpline'', where the phrase ``your call is important to us'' is often repeated, but ``becomes less true over time''. The repair strategies we have proposed here assume the current level of disruption in state-of-the-art HRI, and they may not hold as robot capabilities further develop. \textbf{Future research} should aim to reduce such latency in robotic conversational interaction, so that repair strategies will gradually become less needed.}

\section{Conclusions With A Critical Look}
\label{sec:conclusions}

\hl{
This paper contributes insights for designing repair strategies for longitudinal robotic well-being coaching, informed by real-world users' and a professional coach's perspectives. We have shown how our 4-phase design process and its outcomes can contribute toward the real-world deployment of longitudinal robotic coaches that are capable of repairing their mistakes and thus improving coaching interactions. As part of this process, we designed our initial study to compare \textbf{empathic} and \textbf{non-empathic} repairs, based on two discussions with a professional well-being coach} (Sec. \ref{sec:phase1}). \hl{However, in our between-subjects study where a robotic coach administered these repair strategies} (Sec. \ref{sec:phase2}), \hl{we found that users had detailed feedback 
beyond the question of whether empathic or non-empathic repair strategies were better. We then designed 
a workshop with users to give detailed feedback on repairs for a robotic coach} (Sec. \ref{sec:phase3}), \hl{and reflecting on these user insights with the professional coach from Phase 1} (Sec. \ref{sec:phase4}). 
We encourage researchers to include such retrospective discussions with users and stakeholders in their research, especially when their intuitions are not confirmed, to better understand how robots are experienced in the real world. 




We would also like to direct a critical eye toward our own study. In this paper, we conducted a user study in which the robot made intentional mistakes and repairs. However, \textit{intentional mistakes rely on timing, which is difficult for social robots}. Additionally, the timing of a repair can have an impact on its success \cite{robinette2015timing}. Thoroughly investigating the timing of the mistakes and repairs, and how this may have impacted user perceptions is out of scope of this work. We will investigate this in \textbf{future research}, and encourage the HRI field to reflect on how intentional social mistakes can be investigated taking into account the challenge of timing in interactions.
Also, we used a LLM (ChatGPT) to generate the robot's responses to users. In some cases, the LLM spontaneously asked users for clarification when it did not ``understand'' the user's utterance. In this paper, we asked users to recall the mistakes and repairs they experienced during the interaction (including those spontaneously deployed by the LLM), however, further analysis on spontaneous repairs and users' perceptions of them is out of scope of this work. We invite \textbf{future research} to investigate the impact of \textit{spontaneous} LLM repairs on users' perceptions of robotic coaches.
\hl{These repair strategies have been designed specifically for \textit{well-being coaching}. In previous HRI literature on game- and task-based scenarios, repair strategies focus on a robot's mistakes where a right vs wrong condition is clear due to a set goal} \cite{kontogiorgos2020behavioural, kontogiorgos2020embodiment, kontogiorgos2021systematic, sebo2019don, esterwood2021you, esterwood2023three}. \hl{In conversational contexts such as coaching, identifying the mistake itself, and consequently the necessity for and appropriate type of repair, is more complex due to factors such as user preferences. Despite this complexity, we have attempted to distil relevant insights and reflections on such repair strategies. 
\textbf{Future research} should further investigate longitudinal coaching interactions, administering repair strategies according to our insights, and further refine these in an iterative manner.}







\vspace{-0.2cm}
\begin{acks}
\footnotesize
We thank Cambridge Consultants Inc. and their employees for participating in this study.
\textbf{Funding:} M. Axelsson is funded by the Osk. Huttunen foundation and the EPSRC under grant EP/T517847/1. M. Spitale and H. Gunes have been funded by the EPSRC/UKRI under grant ref. EP/R030782/1 (ARoEQ). \textbf{Open Access:} For open access purposes, the authors have applied a Creative Commons Attribution (CC BY) licence to any Author Accepted Manuscript version arising.
\textbf{Data access:} Raw data related to this publication cannot be openly released due to anonymity and privacy issues.

\end{acks}

\bibliographystyle{ACM-Reference-Format}
\balance
\bibliography{main}

\appendix

\end{document}